\documentclass{article}
\usepackage{spconf,amsmath,graphicx}

\title{Deep Neighbor Layer Aggregation for Lightweight Self-Supervised Monocular Depth Estimation}
\name{Boya Wang$^{\star }$ \qquad Shuo Wang$^{\dagger}$\qquad Dong Ye$^{\star}$ \qquad Ziwen Dou$^{\star}$}
\address{$^{\star}$ Harbin Institute of Technology  \\
      $^{\dagger}$Shanghai Institute of Microsystem and Information Technology}
\begin{document}
%
\maketitle
\begin{abstract}

With the frequent use of self-supervised monocular depth estimation in robotics and autonomous driving, the model's efficiency is becoming increasingly important. Most current approaches apply much larger and more complex networks to improve the precision of depth estimation. Some researchers incorporated Transformer into self-supervised monocular depth estimation to achieve better performance. However, this method leads to high parameters and high computation. We present a fully convolutional depth estimation network using contextual feature fusion. Compared to UNet++ and HRNet, we use high-resolution and low-resolution features to reserve information on small targets and fast-moving objects instead of long-range fusion. We further promote depth estimation results employing lightweight channel attention based on convolution in the decoder stage. Our method reduces the parameters without sacrificing accuracy. Experiments on the KITTI benchmark show that our method can get better results than many large models, such as Monodepth2, with only 30\% parameters. The source code is available at https://github.com/boyagesmile/DNA-Depth.

\end{abstract}
\section{Introduction}
\label{sec:intro}


Depth estimation from a single image is a fundamental and challenging task in computer vision, with applications in 
3D movies, 3D reconstruction, and simultaneous localization and mapping (SLAM). 
However, depth estimation is an ill-posed problem, since different 3D scenes can produce identical 2D images. Traditional methods have relied on exploiting the complex geometric relationship between RGB images and depth maps, but these methods lack flexibility and generality.

Supervised monocular depth estimation methods have recently achieved more accurate results by leveraging prior information. Supervised methods require depth ground truth, which may take a lot of work to get. 
Therefore, improving the performance of supervised learning techniques has become challenging.
Semi-supervised or self-supervised approaches have received much attention to overcome these limitations. Self-supervised stereo depth estimation methods reconstruct depth maps through stereo matching of the left and right views without accurate depth information. Self-supervised methods were proposed to eliminate the dependence on binocular view correction. These approaches use epipolar geometry to construct synthetic perspectives as supervision signals. Self-supervised monocular depth estimation methods still have gaps compared to multi-view and supervised methods.
\begin{figure}[htb]\centering
	\includegraphics[width=8cm]{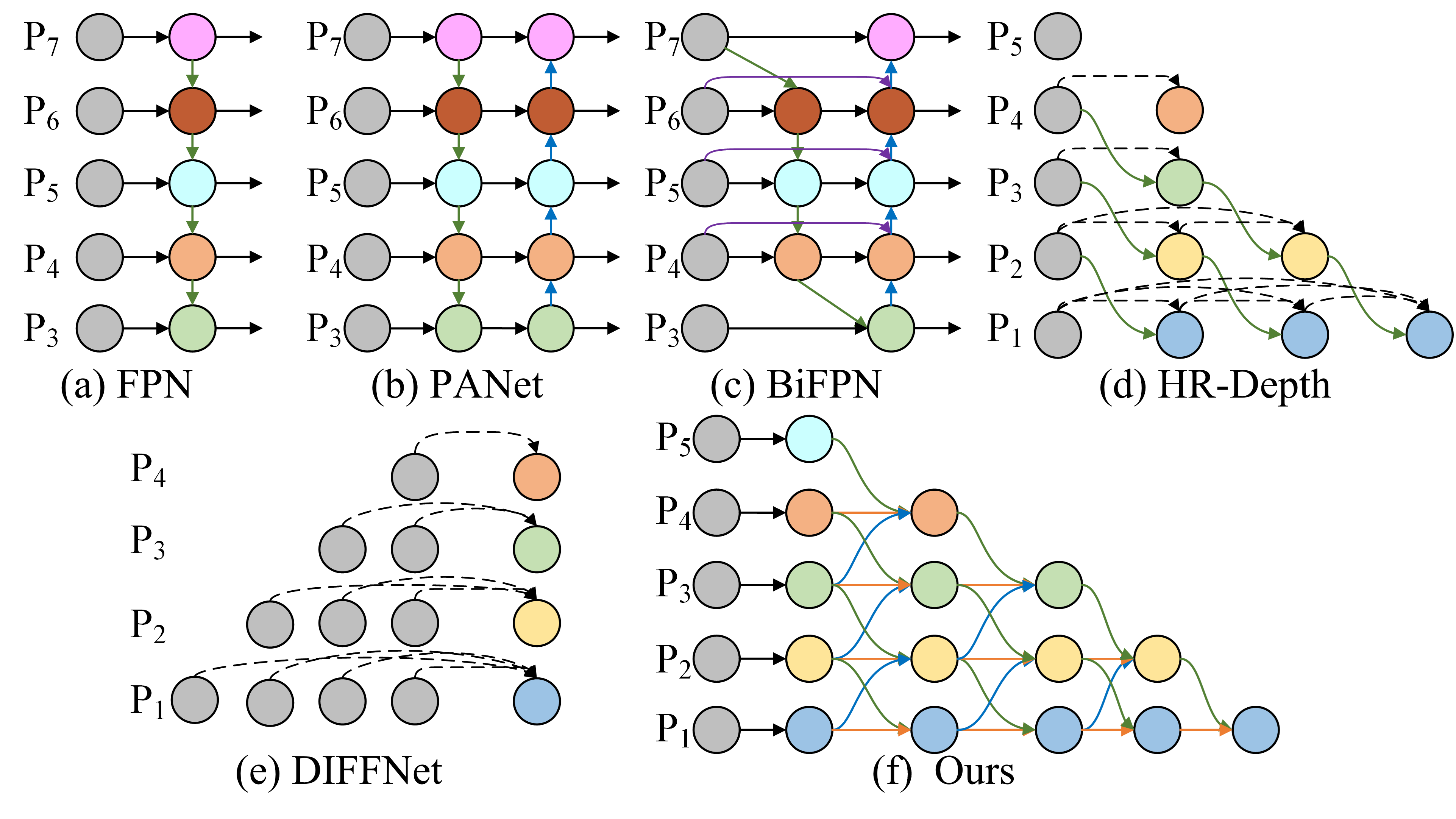}
	\caption{Feature fusion. (a) FPN\cite{lin2017feature} use top-down pathway to fuse multi-scale features,(b) PANet\cite{liu2018path} add down-top pathway based on FPN, (c) BiFPN\cite{tan2019efficientnet}  prune some pathway based on PANet. (d) HRDepth\cite{lyu2021hr} proposes a bottom-up feature fusion and adds skip connections,(e) DiffNet\cite{8962053} fuses the features extracted by the encoder using skip connections. (f) is our contextual feature fusion method to better fuse features of different scales.}\label{fuse}
\end{figure}

When we analyzed the existing methods, we found that most of them are high-parameter models, and the correlation of features is not close enough in the decoding stage. 
This paper introduces a novel approach called DAN-Depth, which leverages contextual feature fusion to enhance the relationship between multi-scale information shown in \mbox{Fig. \ref{fuse}}. We discard long-range connections and only use feature maps of adjacent resolutions for feature fusion.
The main contributions of this paper are: (1) We develop a contextual feature fusion mechanism to fuse multi-stage features and improve the correlation between features. (2) We propose a multi-scale feature focus guide module and focus more on target features to improve the detail of depth estimation. (3) Our proposed method has low parameters but performs better than most depth estimation approaches on the KITTI benchmark. 
\begin{figure*}[htb]\centering
	\includegraphics[width=15cm]{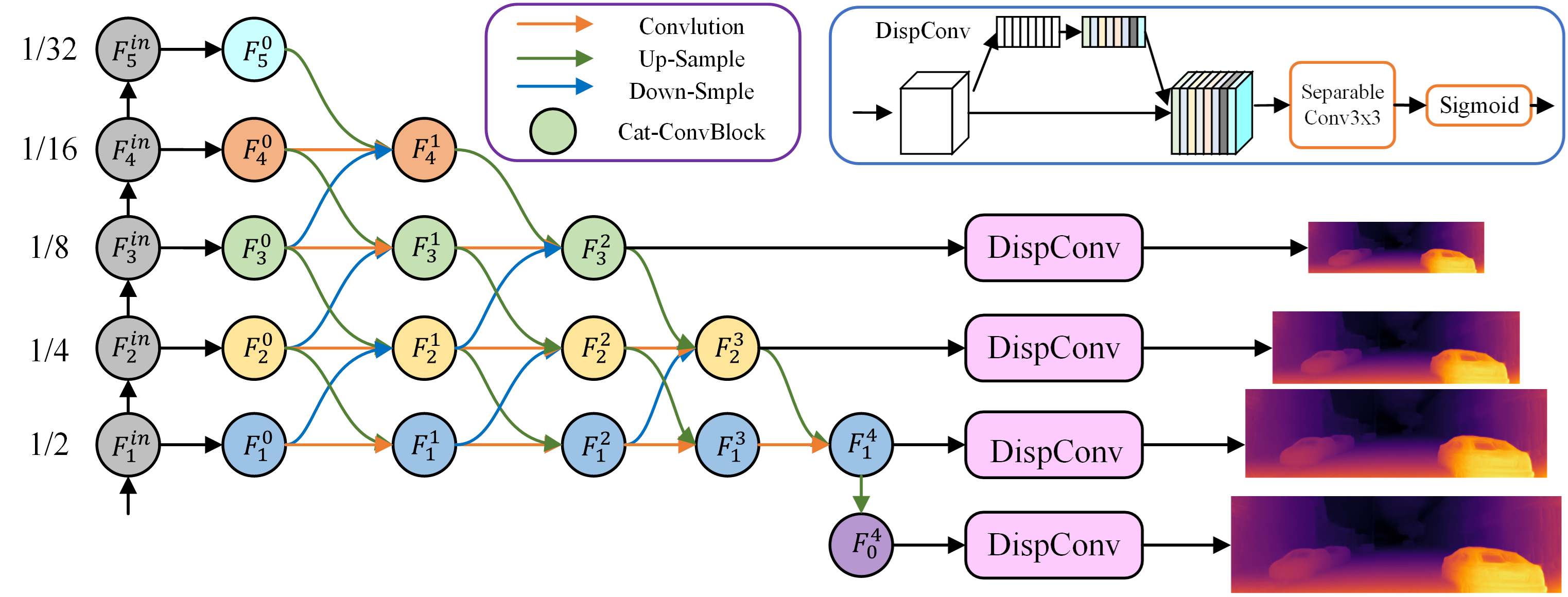}
	\caption{The overview of the DNA-Depth network. EfficientNet is used as the encoder. The decoder employs contextual feature fusion and channel attention to get depth maps of different scales.}\label{network}
\end{figure*}
\section{related work}
\label{sec:format}


Learning-based methods have been shown to solve the ill-posed problem of monocular depth estimation, which can be divided into supervised and self-supervised methods. Several related to the process are reviewed and discussed among the various approaches.

Supervised depth estimation methods require sufficient depth ground-truth for training. Stereo depth estimation networks such as DeMoN\cite{ummenhofer2017demon} and StereoNet\cite{khamis2018stereonet} estimate the depth of the image using stereo image pairs. Supervised monocular depth estimation networks\cite{NIPS2014_7bccfde7} \cite{8658954} only use the input of a single camera to estimate the depth. They used depth, gradients and surface normals to build the loss function to produce higher resolution depth maps.



Self-supervised depth estimation methods, such as Monodepth \cite{Godard_2017_CVPR}, utilize the epipolar geometric constraints between the left and right images and train with an image reconstruction loss to generate disparity maps. However, occlusion can adversely affect the depth estimation results, impacting the value of the loss function. Monodepth2\cite{Godard_2019_ICCV} proposed a novel approach to constructing an anti-occlusion loss function to address this issue. Building upon this work, subsequent studies such as ManyDepth\cite{Watson_2021_CVPR}, HRDepth\cite{lyu2021hr}, and DiffNet\cite{8962053} have further improved the accuracy of monocular depth estimation.


Existing self-supervised methods for monocular depth estimation often need to pay more attention to the relationship between features in the decoding stage. Moreover, these methods typically require complex networks with numerous parameters and computations, which can be resource-intensive and difficult to deploy on devices with limited computing capabilities.

\section{method}
\label{sec:pagestyle}

\subsection{Self-supervised Monocular Depth Estimation}
\label{ssec:subhead}
The self-supervised monocular depth estimation network transforms the depth estimation problem into a new perspective synthesis problem. With the aid of a PoseNet, it predicts the appearance of a target image $I_{t}$ from the viewpoint of a source image $t^{\prime}$.
Extracting interpretable depth is achieved by adding constraints to limit the intermediate variable (depth or disparity).
The DepthNet $f_{Depth}$ takes a target image $I_{t}$ as input and estimates the depth map $D_{t}$. The PoseNet $f_{Pose}$ gets a relative pose change $T_{t \rightarrow t^{\prime}}$ between the target image $I_{t}$ and the source image $t^{\prime}$.
\begin{align}
T_{t \rightarrow t^{\prime}}=f_{Pose}\left(I_{t}, I_{t^{\prime}}\right), t^{\prime} \in\{t-1,t+1\}
\end{align}

The pixels in the source image can be mapped to the target image using the pose $T_{t \rightarrow t^{\prime}}$ and the depth map of the target image $D_{t}$ assuming that the change in image is only caused by the motion of the camera.

\begin{align}
\operatorname{proj}\left(D_{t}, T_{t \rightarrow t^{\prime}}, K\right)=K[T_{t \rightarrow t^{\prime}}(D_{t}K^{-1}P_{t})]
\end{align}

\begin{align}
I_{t^{\prime} \rightarrow t}=I_{t^{\prime}}\left\langle\operatorname{proj}\left(D_{t}, T_{t \rightarrow t^{\prime}}, K\right)\right\rangle 
\end{align}

Here, $D_{t}K^{-1}P_{t}$ project camera points to three-dimensional space, $T_{t \rightarrow t^{\prime}}(D_{t}K^{-1}P_{t})$ move the origin of 3D coordinates to time $t^{\prime}$, $K$ are known as camera intrinsic, $\operatorname{proj}\left(D_{t}, T_{t \rightarrow t^{\prime}}, K\right)$ project the points on 3D coordinates back to 2D coordinates and the $\left\langle\right\rangle $ is the sampling operator.We construct the supervision signal of the network by minimizing the photometric re-project error $re$:
\begin{align}
\begin{split}
re\left(I_{t}, I_{t^{\prime} \rightarrow t}\right)=\frac{\alpha}{2}\left(1-S S I M\left(I_{t}, I_{t^{\prime} \rightarrow t}\right)\right)\\
+(1-\alpha)\left\|I_{t}-I_{t^{\prime} \rightarrow t}\right\|_{1}
\end{split}
\end{align}
where $\alpha=0.85$. The photometric loss is denoted as:
\begin{align}
L_{r e}=\min _{t^{\prime} \rightarrow t} re\left(I_{t}, I_{t^{\prime} \rightarrow t}\right)
\end{align}

Furthermore, we used edge-aware smooth regularization term to normalize differences in textureless low-image gradient regions:
\begin{align}
L_{s m o o t h}=\left|\frac{\nabla D_{t}}{\partial x}\right| e^{-\left|\frac{\nabla I_{t}}{\partial x}\right|}+\left|\frac{\nabla D_{t}}{\partial y}\right| e^{-\left|\frac{\nabla I_{t}}{\partial y}\right|}
\end{align}

So, the ultimate loss function is the sum of the reprojection loss and the smoothing loss:
\begin{align}
L_{f i n a l}=\lambda L_{r e}+ \beta L_{\text {smooth }}
\end{align}
where the $\lambda$ is the weight of the reprojection term and the $\beta$ is the weight of the smooth term.

\subsection{DNA-Depth}
\label{ssec:subhead} 
DNA-Depth proposes a lightweight self-supervised monocular depth estimation method based on contextual feature fusion and channel attention. DNA-Depth is based on an encoding-decoding structure, like most depth estimation networks. \mbox{Fig. \ref{network}} show the network structure.

Existing depth estimation methods often utilize ResNet as an encoder to extract feature maps from input images. However, RestNet has high parameters and computation, which is unsuitable for platforms with limited computing resources. In this paper, we use EfficientNet\cite{tan2019efficientnet} as the encoder with low parameters and computations, but it can obtain accurate spatial features of images. 
EfficientNet was optimized using the neural architecture search method, which balances accuracy and computational efficiency, making it an attractive choice for resource-constrained devices.

We propose a contextual feature fusion method in the design phase of the decoder. We found that input features with a significant difference in scale from the output features have a lower contribution to the output in fully connected feature fusion during our experiments. Therefore, the correlation between features with significant differences in the output feature scale is removed to reduce the number of parameters in the network model. In the feature output process, low-resolution features are gradually discarded and feature aliasing is used to achieve higher-level feature fusion. 
Finally, channel attention is introduced in the output feature map phase to enhance the depth estimation result.

The encoder gives feature maps $\vec{F}^{i n}=\left(F_{l_{1}}^{i n}, F_{l_{2}}^{i n}, \ldots\right)$ at different resolutions, where $F_{l_{i}}^{i n}$ represents the feature maps output by layer $l_{i}$. The feature extraction network $\mathcal{E}(\cdot)$ is a feature extraction block, which is used to get $F_{i,0}$ by unifying the channel numbers of feature graphs of different dimensions.
\begin{align}
F_{i,0}=\mathcal{E}(F_{i})
\end{align}

We add intermediate nodes to the decoder to perform contextual feature aggregation. We assume that $F_{{i}}^{j}$ is the output of the current node, where $i$ denotes the resolution of the feature and $j$ denotes the feature of different stages. The stack of feature maps is calculated as
\begin{align}
F_i^j=\left\{\begin{array}{ll}\mathcal{C}(\left[\mathcal{U}(F_{i+1}^{j-1}),\mathcal{S}(F_{i}^{j-1})\right]),&i=1\\\mathcal{C}(\left[\mathcal{U}(F_{i+1}^{j-1}),\mathcal{S}(F_{i}^{j-1},\mathcal{D}(F_{i-1}^{j-1})\right]),&i>1\end{array}\right.
\end{align}
where $\mathcal{S}(\cdot)$ is a separable convolution block, $\mathcal{U}(\cdot)$ is an upsample block, $\mathcal{D}(\cdot)$ is a downsample block. $\left[\cdot\right]$ is a concatenation layer. $\mathcal{C}(\cdot)$ denotes a feature fusion block consisting of a convolution operation and an activate function.

To improve the accuracy of the output depth map, channel attention is used in the output part. Let $F_i^{out}$ denote the last stage of the feature. The output depth map can be expressed as:
\begin{align}
D_i=\mathcal{S}_i(\mathcal{S}_{conv}(\mathcal{CA}(F_i^{out})))
\end{align}

where $\mathcal{CA}(\cdot)$ is a channel attention module, $\mathcal{S}_{conv}(\cdot)$ denotes a separable convolution block, $\mathcal{S}_i(\cdot)$ denotes a sigmoid function.

\begin{table*}[!t]
    \caption{\label{table1}The KITTI benchmark test results. The 640M uses 640$\times$192 images for training, while the 1024M uses 1024$\times$320 images. The marked with bold lines from the method presented in this paper.}
    \centering
    
    \begin{tabular}{ccccccccccc}
    \hline\\[-3mm]
Method             & train & Abs Rel & Sq Rel & RMSE  & RMSElog &  $\delta$1    & $\delta$2    & $\delta$3    & FLops& params \\
    \hline\\[-3mm]
Monodepth2         & 640M  & 0.115   & 0.903  & 4.863 & 0.193   & 0.877 & 0.959 & 0.981 & 8.04        & 14.84      \\
HRDpeth            & 640M  & 0.109   & 0.792  & 4.632 & 0.185   & 0.884 & 0.962 & 0.983 & 18.08       & 14.61      \\
DIFFNet            & 640M  & 0.102   & 0.764  & 4.483 & 0.180   & 0.896 & 0.965 & 0.983 & 15.77       & 10.87      \\
{ \bf DNA-Depth-B0} & 640M  & { \bf0.105}   & { \bf0.748}  & { \bf4.489} & { \bf0.179}   & { \bf0.892 }& { \bf0.965 }& { \bf0.984} & { \bf9.87  }      & { \bf4.15}       \\
{ \bf DNA-Depth-B1}  & 640M  & { \bf0.102 }  & { \bf0.757 } &{ \bf4.493 }& { \bf0.178}   &{ \bf 0.896} &{ \bf 0.965 }& { \bf0.984} & { \bf10.35 }      & { \bf6.66}       \\
    \hline\\[-3mm]
Monodepth2         & 1024M  & 0.115   & 0.882  & 4.701 & 0.190   & 0.879 & 0.961 & 0.982 &     21.45        &    14.84        \\
HRDpeth            & 1024M  & 0.106   & 0.755  & 4.472 & 0.181   & 0.892 & 0.966 & 0.984 &      48.22&14.61            \\
DIFFNet            & 1024M  & 0.097   & 0.722  & 4.345 & 0.174   & 0.907 & 0.967 & 0.984 &      42.06       &       10.87     \\
{ \bf DNA-Depth-B0} & 1024M  & { \bf0.100 }  & { \bf0.703}  & { \bf4.344} & { \bf0.174 }  & { \bf0.898} & { \bf0.967} & { \bf0.985} &      { \bf26.35 }      &   { \bf 4.15  }      \\
{ \bf DNA-Depth-B1}  & 1024M  & { \bf0.097}   & { \bf0.682}  & { \bf4.357} & { \bf0.174}   & { \bf0.902} & { \bf0.968} & { \bf0.984} &     { \bf 27.58 }      &     { \bf 6.66}     \\
    \hline\\[-3mm]
    \end{tabular}
\end{table*}

\begin{figure*}[htb]\centering
	\includegraphics[width=17cm]{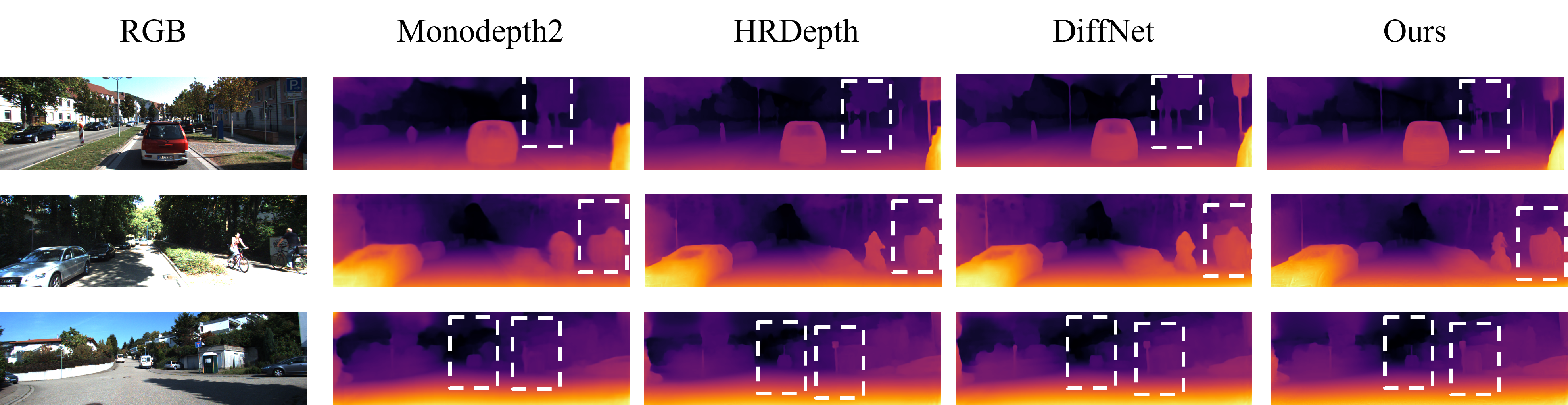}
	\caption{Depth estimation results visualization. The left column contains the input RGB images. The right column shows the result from DNA-Depth-B1; the remaining columns are from other contemporary methods. The results show that our network can predict small objects and moving objects with comparable performance to DIFFNet but far fewer parameters.
 }\label{compare}
\end{figure*}

\section{experiments}
\label{sec:typestyle}


\subsection{KITTI Datasets}
KITTI \cite{kitti} is a dataset for evaluating the computer vision algorithm in autonomous driving scenarios, including two greyscale cameras, two color cameras, and a 64-line Velodyne LiDAR. 
We adopted the dataset split method suggested by \cite{NIPS2014_7bccfde7} and removed the static frames proposed by \cite{Zhou_2017_CVPR}. Finally, we obtained 39810 frames for training, 4424 for verification, and 697 for evaluating model performance. Also, to simplify the training process, we assume that the camera intrinsic is the same in all scenarios. 

\subsection{Implementation Details}

Our model is implemented in Pytorch framework\cite{NEURIPS2019_bdbca288} and trained on a single NVIDIA A100 GPU. We use the Adam Optimizer\cite{kingma2014adam} with default betas 0.9 and 0.999. The DepthNet and PoseNet are trained for 40 epochs with a batch size of 16. We set the initial learning rate to  $10^{-4}$ for the first 35 epochs and then drop to $10^{-5}$ for fine-tuning the remainder. We set the SSIM weight to $\alpha=0.85$ and the smoothing loss term weight to $\lambda =10^{-3}$.

We utilize EfficientNet-B0 and EfficientNet-B1 as the backbone architectures for our model. To initialize the DNA-Depth model, we employ the pre-trained EfficientNet model. We compute the average loss across four scaled depth maps during training. The model generates a single high-resolution depth map at inference time as output.
Regarding the PoseNet, it is built upon the ResNet-18 architecture. Given two consecutive video frames as input, the PoseNet produces a 6-DOF vector representing the relative pose between the two frames.

\subsection{Evaluation on KITTI Benchmark}
We assessed the performance of DNA-Depth on the KITTI benchmark using the evaluation metrics presented in \cite{NIPS2014_7bccfde7}. The quantitative results are shown in Table \ref{table1}. Our method demonstrates a favorable trade-off between computational efficiency and accuracy, achieving high performance across various indices while maintaining a low parameter count and computational complexity. In particular, DNA-Depth yields improved accuracy when taking 1024 × 320 resolution images as input, outperforming other state-of-the-art methods. Fig. \ref{compare} provides a visual comparison of the qualitative properties of DNA-Depth against DIFFNet, HR depth, and Monodepth2.

\section{conclusion}
\label{sec:majhead}

This study introduces a novel approach to self-supervised monocular depth estimation called DNA-Depth. Leveraging the efficacy of EfficientNet as an encoder, we focus our innovations on the decoding stage. Specifically, we propose a fully convolutional depth estimation network that harnesses contextual feature fusion during the feature fusion stage. We generate high-quality depth maps with enhanced accuracy by fusion of contextual features and channel attention. Our method undergoes evaluation on the KITTI benchmark, demonstrating superior performance while simultaneously boasting reduced computational complexity and parameter count compared to previous state-of-the-art approaches.

\bibliographystyle{IEEEbib}
\bibliography{strings,refs}

\end{document}